\documentclass[10pt,twocolumn,letterpaper]{article}

\usepackage[pagenumbers]{iccv} 

\definecolor{iccvblue}{rgb}{0.21,0.49,0.74}
\usepackage[pagebackref,breaklinks,colorlinks,allcolors=iccvblue]{hyperref}

\def\paperID{14898} 
\def\confName{ICCV}
\def\confYear{2025}

\title{MAGIC-Talk: Motion-aware Audio-Driven Talking Face Generation with 
Customizable Identity Control}

\author{
Fatemeh Nazarieh$^{1}$,
Zhenhua Feng$^{2}$,
Diptesh Kanojia$^{1}$,
Muhammad Awais$^{3}$,
and Josef Kittler$^{3}$\\[4pt]
$^{1}$School of Computer Science and Electronic Engineering, University of Surrey\\
$^{2}$School of Artificial Intelligence and Computer Science, Jiangnan University\\
$^{3}$Centre for Vision, Speech and Signal Processing, University of Surrey\\[4pt]
{\tt\small \{f.nazarieh\}@surrey.ac.uk}
}

\begin{document}
\maketitle
\begin{abstract}
Audio-driven talking face generation has gained significant attention for applications in digital media and virtual avatars. While recent methods improve audio-lip synchronization, they often struggle with temporal consistency, identity preservation, and customization, especially in long video generation. To address these issues, we propose MAGIC-Talk, a one-shot diffusion-based framework for customizable and temporally stable talking face generation. MAGIC-Talk consists of ReferenceNet, which preserves identity and enables fine-grained facial editing via text prompts, and AnimateNet, which enhances motion coherence using structured motion priors. Unlike previous methods requiring multiple reference images or fine-tuning, MAGIC-Talk maintains identity from a single image while ensuring smooth transitions across frames. Additionally, a progressive latent fusion strategy is introduced to improve long-form video quality by reducing motion inconsistencies and flickering. Extensive experiments demonstrate that MAGIC-Talk outperforms state-of-the-art methods in visual quality, identity preservation, and synchronization accuracy, offering a robust solution for talking face generation.
\end{abstract}

\section{Introduction}
\label{sec:intro}

\begin{figure*}[!t]
    \centering
    \includegraphics[trim={1cm 14.3cm 1cm 1cm}, clip, width=\textwidth]{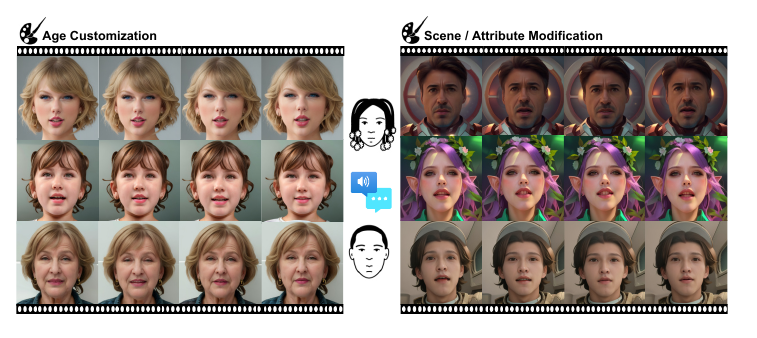}
    \caption{Illustration of our proposed MAGIC-Talk framework for customizable and temporally consistent talking face generation. Given a single reference image, speech audio, and a text prompt, our model enables fine-grained control for talking face generation.}
    \label{fig0}
\end{figure*}

Audio-driven talking face generation animates a static portrait using speech audio. It has gained significant attention for applications such as virtual avatars, filmmaking, gaming, and digital content creation~\cite{10385218}.  
Early approaches~\cite{Chung16a,wav2lip} focused on mapping speech audio to lip movements but often resulted in rigid and unrealistic animations, as only the mouth was animated while the rest of the face remained static. Later approaches~\cite{makeittalk,Zhang_2023_CVPR,wang2021audio2head} attempted to introduce full-face motion, but their expressiveness remained limited due to constraints in the generative capacity. 
Recent advancements in video diffusion models~\cite{xu2024hallohierarchicalaudiodrivenvisual,wei2024aniportrait,2024portraittalk,tian2024emoemoteportraitalive} significantly improved the realism of audio-driven talking face generation. Existing diffusion-based approaches~\cite{stableTalk,Shen_2023_CVPR,tian2024emoemoteportraitalive} integrate concatenated audio and reference image features through a shared attention mechanism to guide the video synthesis. 
Further, to improve motion smoothness, these methods often adopt an autoregressive strategy, where the frames are generated sequentially, based on those synthesized previously. However, these techniques face key challenges. Concatenating audio and reference frame limits audio-visual understanding, while the conditioning on a small set of past frames can introduce temporal drift.

To incorporate emotion into generated talking faces, existing methods use either a single emotion label~\cite{EmoGen,Tan_2024_CVPR} or an emotion reference video~\cite{EAMM,edtalk} to guide facial expressions during generation. However, assigning emotion labels and using fixed expression templates can not capture the subtle emotional variations naturally present in speech, leading to inconsistent facial expressions. 
Beyond emotional control, most talking face generation methods~\cite{makeittalk,wang2021audio2head,stableTalk,stypulkowski2024diffused} rely on audio as the main conditioning source for generation, with a limited exploration of text-based control. Although text prompts provide a great flexibility for customizable generation, existing text-driven methods~\cite{Song_2022_CVPR,2024portraittalk,li2021writeaspeaker,tan2024style2talker,Jang_2024_CVPR} often restrict modifications to specific facial regions or styles, resulting in unacceptable identity preservation and limited overall controllability.

To address these challenges in talking face generation, we propose MAGIC-Talk, a one-shot, \textbf{M}otion-\textbf{A}ware and \textbf{G}eneralizable \textbf{I}dentity-Preserving \textbf{C}ustomized diffusion-based framework designed to ensure identity consistency, temporal stability, and customizability, while generating high-fidelity \textbf{talk}ing face videos from a single reference image, speech audio, and textual description. 
Our framework consists of two key components: ReferenceNet and AnimateNet. \textbf{ReferenceNet} integrates an appearance encoder to extract rich identity-specific features from the reference image, enhancing identity preservation beyond CLIP-based~\cite{ye2023ipAdapter,li2023photomaker} approaches. Additionally, a decoupled cross-attention mechanism processes identity and non-identity-related features separately, preventing identity drift, while allowing fine-grained facial attribute customization based on the user-provided text descriptions. A significant challenge in talking face generation is temporal consistency, as motion inconsistencies often lead to flickering and unnatural transitions. To address this, we incorporate motion modules into ReferenceNet to model realistic facial dynamics, including head movements and eye blinks. While the motion module improves overall video dynamics, it does not inherently ensure accurate audio-lip synchronization. 
To bridge this gap, we utilize a pre-trained variational motion generator~\cite{ye2023geneface++} to map audio features to the corresponding facial landmarks, ensuring precise alignment between speech and facial motion. \textbf{AnimateNet} then leverages these extracted motion priors to achieve precise audio-lip synchronization. 

While these components ensure identity preservation and synchronized motion in shorter clips, generating long-form videos presents additional challenges, including maintaining consistency over extended sequences. To address this, we introduce a training-free progressive sampling fusion strategy, which processes video in overlapping temporal segments. 
By progressively refining the motion representation at each step, our approach effectively extends video length, while maintaining identity stability and motion coherence. Our framework offers a robust one-shot solution for customizable and temporally consistent talking face generation, with applications in virtual avatars, filmmaking, digital content creation, and interactive media.

In summary, our contributions are threefold. (1) We propose MAGIC-Talk, a novel one-shot diffusion-based talking face generation framework that integrates precise appearance encoding and text prompts to guide the generation pipeline toward customizable and generalizable talking face synthesis, while ensuring temporal consistency and accurate audio-lip synchronization. (2) To support long video generation, MAGIC-talk incorporates a progressive sampling fusion strategy that processes video in overlapping segments, ensuring smooth transitions, mitigating motion inconsistencies, and preventing temporal drift. (3) The results of qualitative and quantitative analysis demonstrate that MAGIC-Talk outperforms state-of-the-art methods in identity preservation, motion realism, and synchronization accuracy across diverse identities and textual descriptions.

\section{Related Works}
\label{relatedworks}
\subsection{Audio-driven Talking Face Generation}
Audio-driven talking face generation focuses on synthesizing talking face videos using only audio as input. Early works primarily focused on synchronization of lip movements with the driving speech signal. For instance, Chung et al.~\cite{Chung16a} introduced an encoder-decoder model for lip movement generation. While their approach laid the foundation for the task, this particular  method limited motion primarily to the mouth region, resulting in relatively static videos. The subsequent efforts aimed to enhance naturalness by leveraging intermediate representations such as facial landmarks~\cite{makeittalk} and dense motion fields~\cite{wang2021audio2head}. 
Notable methods such as MakeItTalk~\cite{makeittalk} and SadTalker~\cite{Zhang_2023_CVPR} employed intermediate features to guide facial animation, while others~\cite{Peng_2023_ICCV,Tan_2024_CVPR,Peng_2024_CVPR} incorporated 3D information to improve head movements and overall realism. 
Despite these advancements, the generated faces often suffered from distortions, inconsistent identity features, and lack of emotional control.

To address facial expressiveness, several approaches~\cite{EmoGen,Tan_2024_CVPR} incorporated one-hot vectors representing predefined emotions to generate emotional talking face videos. While this enabled some degree of emotional control, the reliance on discrete emotion labels constrained the diversity of expressions. Other methods, such as EAMM~\cite{EAMM} and EDTalk~\cite{edtalk}, transferred expressions from an emotional source video to the target speaker, enhancing expressiveness and head movements. However, these approaches frequently encountered irregularities and audio-lip synchronization issues, especially when dealing with unseen characters or audio inputs. Recently, diffusion-based models~\cite{wei2024aniportrait,xu2024hallohierarchicalaudiodrivenvisual,2024portraittalk} 
have demonstrated notable improvements in audio-lip synchronization. Nonetheless, these models face challenges such as identity inconsistencies, visual artifacts with new identities, and a limited ability to customize the video’s style or content based on user descriptions.

\subsection{Text-to-Video Generation}
Recent advancements in large text-to-image models~\cite{Singer2022MakeAVideoTG,zhou2023magicvideoefficientvideogeneration,esser2023structurecontentguidedvideosynthesis,wang2023videocomposercompositionalvideosynthesis} have enabled the synthesis of diverse, high-fidelity images from text prompts. However, extending these capabilities to video generation presents greater challenges, including maintaining temporal coherence and controlling motion dynamics across frames. Recent progress in diffusion-based video generation~\cite{esser2023structurecontentguidedvideosynthesis,wang2023videocomposercompositionalvideosynthesis,Singer2022MakeAVideoTG} has shown promising results, leveraging foundational principles from text-to-image diffusion models. One of the pioneering works in this area is the Video Diffusion Model (VDM~\cite{SDiffusion}), which introduced a space-time factorized UNet for video generation. While novel, the generated videos often exhibit poor visual quality and severe artifacts. Subsequently, models like Make-A-Video~\cite{Singer2022MakeAVideoTG} and Magic Video~\cite{zhou2023magicvideoefficientvideogeneration} advanced text-to-video generation but lacked mechanisms for fine-grained control over the appearance and motion of the generated content. 
To address this limitation, later works explored conditional diffusion processes by integrating structure-guided elements. For instance, Gen-1~\cite{esser2023structurecontentguidedvideosynthesis} and Video Composer~\cite{wang2023videocomposercompositionalvideosynthesis} are among the first methods to employ structural guidance for enhanced video generation. Although general-domain text-to-video methods have shown encouraging results, their applicability to audio-driven talking face generation is marred by the lack of alignment, identity preservation, and motion control. 
\begin{figure*}[!t]
    \centering
    \includegraphics[trim={1.2cm 17.2cm 3.4cm 1cm}, clip, width=.9\textwidth]{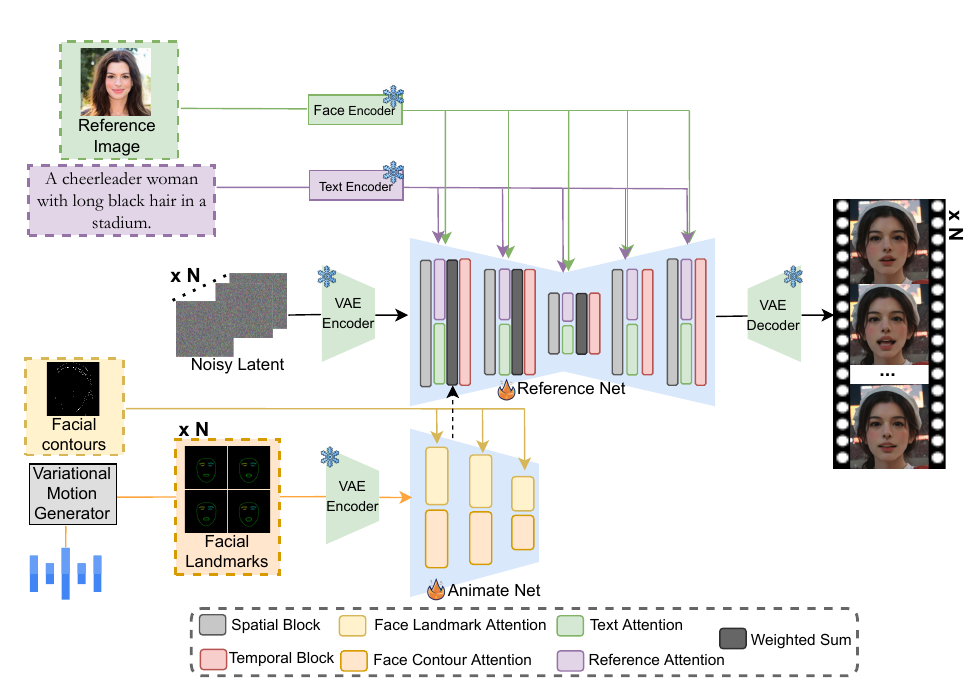}
    \caption{An overview of proposed MAGIC-Talk framework for one-shot, customizable talking face generation. 
    The framework consists of two key components: ReferenceNet, which preserves identity, while enabling fine-grain facial editing through text guidance, and AnimateNet, which maps structured motion priors to enhance temporal coherence and speech-driven dynamics.
    }
    \label{fig2}
\end{figure*}

\section{Methodology}
\label{method}
Given a single reference image, speech audio, and text description, MAGIC-Talk generates customizable talking face videos while preserving identity and ensuring accurate audio-lip synchronization. As shown in Figure~\ref{fig2}, our framework comprises two main components: ReferenceNet and AnimateNet. The following sections detail each component.

\subsection{RefrenceNet}
The core objective of our ReferenceNet is to generate a customized talking face for a specific identity, based on a given reference image and a text prompt. To achieve identity preservation and customization, we move beyond traditional feature concatenation approaches~\cite{Shen_2023_CVPR,stypulkowski2024diffused,stableTalk}, which are often insufficient to capture essential facial details for realistic and consistent talking face generation. Instead, we adopt a decoupled cross-attention mechanism~\cite{ye2023ipAdapter}, where separate cross-attention layers are added to the original UNet architecture. This design allows independent processing of image and text features, with the final feature vector obtained by summing the outputs of these layers for effective fusion. 

Specifically, a pre-trained face encoder~\cite{wang2024instantid}, extracts features from the reference image to guide identity-preserved personalization. Alongside identity preservation, customization is achieved through text descriptions processed with the CLIP text encoder. Text and image embeddings are handled separately in their respective cross-attention layers and summed to serve as the input for the subsequent layers, enabling both identity preservation and text-based manipulation. 
To enhance temporal consistency and natural facial movements, we incorporate fine-tuned motion blocks~\cite{guo2023animatediff}, placed between 2D layers (Section~\ref{MotionB}) to facilitate cross-frame information exchange. The training objective of ReferenceNet mirrors that of image-based generative models by predicting the noise added to latent features ($z^{1:N}$) over $N$ frames and minimizing the error using the following loss function:
\begin{equation}
    \text{loss}_{train} = \mathbb{E}_{t,\mathbf{z}^{1:N}, c, \epsilon \sim \mathcal{N}(0,1)}\left[ \| \epsilon - \epsilon_\theta(\mathbf{z}^{1:N}, t, \mathbf{c}) \|^2 \right]
\end{equation}
where $t$ is the diffusion steps and $c_{t}$ is the condition set (text and image). 

\subsection{AnimateNet}
\label{animateNet}
Mapping audio directly to its corresponding lip movements in a talking face video is a challenging task due to the inherent differences between audio and visual modalities. An effective approach to bridge this gap is to first map audio to motion, and then transfer these motion priors to the visual domain. 
Following~\cite{2024portraittalk}, we integrate a pre-trained Variational Motion Generator~\cite{ye2023geneface++} into our framework. This module employs the HuBERT~\cite{HuBERT} audio transformer to extract phoneme-aware speech embeddings, which are then mapped to expressive facial motion. Motion priors are derived by measuring the deviation of key points on a 3D Morphable Model (3DMM) from the mean mesh, effectively capturing facial dynamics. These priors are then mapped to the corresponding video frames for talking face synthesis. 

To achieve this, we propose AnimateNet, which extends a pre-trained diffusion model following a ControlNet-inspired~\cite{controlnet} design. AnimateNet incorporates a cloned network with trainable control layers and ZeroConv layers allowing the integration of motion priors while maintaining the base model’s generative capacity. For visualization (Figure~\ref{fig2}), we illustrate only the encoder part of AnimateNet, highlighting the trainable control layers and their interaction with the other components of the framework. 

While facial landmarks improve audio-lip synchronization, relying solely on this condition can lead to facial distortions and diminished realism in the generated output. This issue is further amplified, particularly in our one-shot setting where the model must also infer facial structure and identity-specific attributes from just a single reference image. To address this, we incorporate image contours as an additional conditioning signal using an edge detection model, specifically Canny edge detection. Image contours capture essential structural information to guide the layout of the generated talking face. This condition can be extracted from either the reference image or a user-specified image but must remain unchanged throughout the generation process to ensure stable and coherent facial synthesis. 

To improve feature integration, we adopt a decoupled cross-attention mechanism, as in ReferenceNet, to process each condition independently. This mechanism enhances the model controllability, while ensuring a smooth fusion of the features within the generative network. The final output, $Z_{\text{new}}$, is computed as a weighted sum of all attention blocks and serves as input to the subsequent layers. This computation is defined as follows:
\begin{align}
 \label{eq2}
   Z_{\text{new}}^{1:N} = & \, w_{1} \left( \text{CrossAttn}_{\textbf{Face-Landmark}}(Q,K_{1},V_{1}) \right) \nonumber \\
   & + w_{2} \left( \text{CrossAttn}_{\textbf{Face-Contour}}(Q,K_{2},V_{2}) \right) 
\end{align}
The attention score is computed using $\text{CrossAttn}(Q,K_{i},V_{i})$, following the standard attention mechanism~\cite{NIPS2017_3f5ee243}. Here, $Q$, $K$, and $V$ correspond to the query, key, and value matrices, respectively. The key and value matrices are independently computed for each condition set, while the query is shared across all attention blocks. Weights $w_1$ and $w_2$ are assigned to each attention block and initialized equally during training to ensure balanced importance. 
By conditioning on both motion priors and image contours, our framework animates the reference identity while preserving facial structure, maintaining identity consistency, and ensuring smooth transitions between frames. 
Notably, the processed information from AnimateNet is integrated into ReferenceNet through a weighted sum of attention blocks, ensuring a cohesive and controlled synthesis process.

\subsection{Motion Block}
\label{MotionB}
Temporal smoothness is a critical aspect of audio-to-talking face generation. To achieve this, we developed our motion blocks based on~\cite{guo2023animatediff} and incorporated each block after the spatial blocks in the ReferenceNet. 
These motion blocks utilize a temporal attention mechanism with position encoding, which captures the relationships between the consecutive frames in talking-face videos. Positional encoding plays a crucial role in making the model aware of each frame's position within the video. To be specific, the original 2D UNet is inflated into a 3D temporal UNet by integrating motion blocks into our model. The randomly initialized latent noise with $b$ batch size, $c$ channel, $h$, $w$ spatial details and $N$ number of frames ($z_{t}^{1:N} \in \mathbb{R}^{b \times c \times N \times h \times w}$) is reshaped to $\mathbb{R}^{(b \times N) \times c \times h \times w}$. It serves as the primary input to the generative model. Within the motion blocks, the features are reshaped again, this time to $\mathbb{R}^{(b \times h \times w) \times N \times c}$, to process each frame independently while facilitating cross-frame information exchange through the subsequent temporal attention mechanism. 
The temporal attention mechanism follows the standard attention~\cite{NIPS2017_3f5ee243} operation. It is computed as: 
\begin{align}
    \text{attention}_{temporal} = \text{softmax}\left(\frac{Q \cdot K^\top}{\sqrt{d_k}}\right)V
\end{align}
where $Q$, $K$ and $V$ are query, key and value matrices and $d_k$ is the key's dimension. 
Through this attention mechanism, ReferenceNet aggregates temporal information from neighboring frames, synthesizing $N$ frames with improved temporal consistency. Once the motion module processes the frames, the original spatial dimensions are restored by reshaping the tensors to $\mathbb{R}^{(b \times N) \times c \times h \times w}$, ensuring seamless integration between temporal and spatial features.

\subsection{Long-form Video Generation}
While current video generation models~\cite{guo2023animatediff,Zhang_2024_CVPR,Chen_2024_CVPR} exhibit impressive capabilities, they are constrained to generating videos with a fixed number of frames. This limitation arises from the computational complexity of temporal attention, which scales quadratically with the number of frames, making the generation of extended videos computationally expensive. 
Recent research~\cite{stypulkowski2024diffused, Shen_2023_CVPR, stableTalk} explored autoregressive approach to mitigate computational complexity in sequential long video generation. However, these approaches often degrade quality and disrupt temporal consistency~\cite{mimicmotion2024}. To address these issues, we draw inspiration from~\cite{wang2023gen,mimicmotion2024} and introduce a progressive sampling fusion strategy. 
Progressive sampling fusion is a training-free technique integrated into the denoising process of the latent diffusion model during inference. It partitions a long motion sequence into fixed-length segments of $N$ frames with an overlap of $C$ frames ($C > 0$), ensuring smooth transitions and frame-wise coherence. 

At each denoising step $t$, video segment $i$ is processed independently while conditioned on the same reference image, text prompt, and corresponding motion priors. 
The latent representation at each timestep $t$ is updated using a weighted interpolation:
\begin{equation}
x_C^t = \alpha_j x_C^{t, (i)} + (1 - \alpha_j) x_C^{t, (i+1)} 
\end{equation}
where $x_C^{t, (i)}$ and $x_C^{t, (i+1)}$ are the corresponding latents for the overlapping frames from adjacent segments at denoising step $t$. The blending coefficient $\alpha$ is defined as:
\begin{equation} 
\alpha_j = \frac{j}{C}, \quad j \in [0, C] 
\end{equation}
where $j$ denotes the frame index within the overlapping region. If $j=0$, then $\alpha_0=0$, meaning the frame is fully influenced by the previous segment. Conversely, if $j=C$, then $\alpha_C = 1$, making the frame entirely determined by the next segment. This weighting scheme ensures a gradual transition between segments, preserving temporal consistency while preventing abrupt changes and flickering artifacts. 
The final latent representation is decoded into video frames via the diffusion decoder. 
In our work, we found that a segment length of $16$ frames with an $8$ frame overlap yielded the best balance of quality and temporal consistency. However, optimal settings may vary based on model architecture and motion complexity.

\section{Datasets and Evaluation Metrics}

We fine-tune our framework on HDTF~\cite{zhang2021flow} and MEAD~\cite{kaisiyuan2020mead} datasets, two widely used benchmarks for audio-to-talking face generation. Since these datasets lack text prompts, we manually generate descriptive prompts by including key information about each corresponding frame. 
To comprehensively assess our framework, we employ widely used metrics in talking face generation. Specifically, PSNR~\cite{PSNR}, SSIM~\cite{SSIM}, and FID~\cite{NIPS2017_8a1d6947} measure visual fidelity, while Landmark Distance (LMD)~\cite{LMD} evaluates facial landmark accuracy on both face and mouth regions.  
SyncNet is used to assess audio-lip synchronization and temporal consistency. Additionally, to evaluate the impact of customization in generated talking faces, we include CLIP-T\cite{pmlrv139radford21a} to measure prompt fidelity, and DINO\cite{Caron_2021_ICCV} and Face similarity~\cite{li2023photomaker} to assess identity consistency. For details on implementation and datasets, please refer to the supplementary material.

\begin{figure}[!t]
    \centering
    \includegraphics[trim={1cm 7cm 1cm 1cm}, clip, width=\columnwidth]{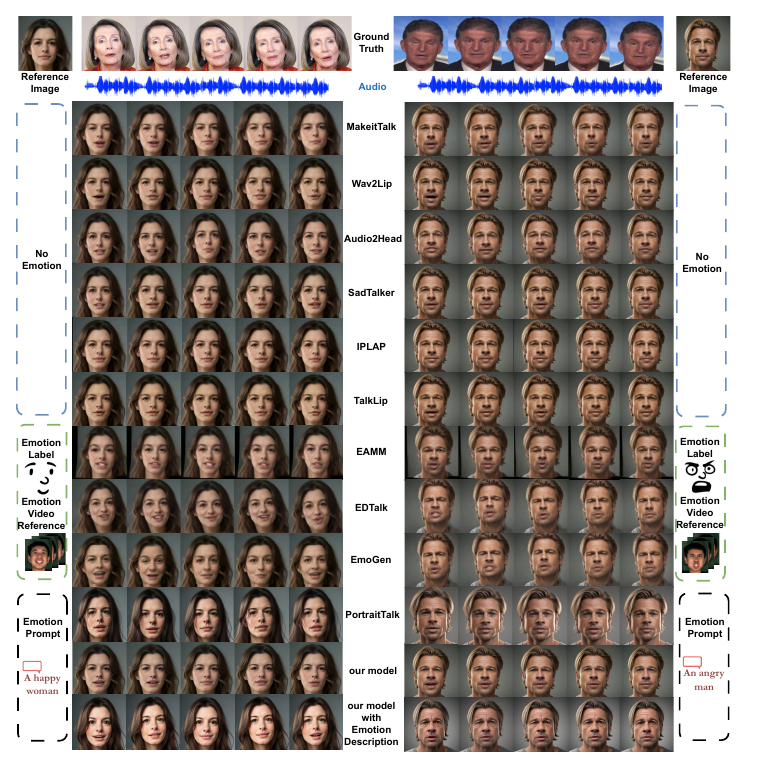}
    \caption{Qualitative comparison of our method with baseline talking face generation approaches. The methods are categorized into three groups: (1) No emotion conditioning, (2) Emotion label or reference video guidance, and (3) Text description guidance.
    }
    \label{fig3}
\end{figure}

\section{Results}

\begin{table*}[ht]
\centering

\label{table1}
\resizebox{\textwidth}{!}{%
\begin{tabular}{lccccccccccccc}
\toprule
\textbf{Method} & \multicolumn{5}{c}{\textbf{MEAD}~\cite{kaisiyuan2020mead}} & \multicolumn{5}{c}{\textbf{HDTF}~\cite{zhang2021flow}} \\
\cmidrule(lr){2-6} \cmidrule(lr){7-11}
& \textbf{PSNR↑} & \textbf{SSIM↑} & \textbf{M/F-LMD↓} & \textbf{FID↓} & \textbf{SyncNet ↑} & \textbf{PSNR↑} & \textbf{SSIM↑} & \textbf{M/F-LMD↓} & \textbf{FID↓} & \textbf{SyncNet ↑}\\
\midrule
MakeItTalk~\cite{makeittalk} & 19.442 & 0.614 & 2.541/2.309 & 37.917 & 5.176 & 21.985 & 0.709 & 2.395/2.182 & 18.730 & 4.753 \\
Wav2Lip~\cite{wav2lip}    & 19.875 & 0.633 & 1.438/2.138 & 44.510 & 8.774 & 22.323 & 0.727 & 1.759/2.002 & 22.397 & \textbf{9.032}  \\
Audio2Head~\cite{wang2021audio2head} & 18.764 & 0.586 & 2.053/2.293 & 27.236 & 6.494 & 21.608 & 0.702 & 1.983/2.060 & 29.385 & 7.076 \\
SadTalker~\cite{Zhang_2023_CVPR}  & 19.042 & 0.606 & 2.038/2.335 & 39.308 & 7.065 & 21.701 & 0.702 & 1.995/2.147 & 14.261 & 7.414 \\
IP-LAP~\cite{Zhong_2023_CVPR}     & 19.832 & 0.627 & 2.140/2.116 & 46.502 & 4.156 & 22.615 & 0.731 & 1.951/1.938 & 19.281 & 3.456 \\
TalkLip~\cite{wang2023seeing}    & 19.492 & 0.623 & 1.951/2.204 & 41.066 & 5.724 & 22.241 & 0.730 & 1.976/1.937 & 23.850 & 1.076  \\
EAMM~\cite{EAMM}  & 18.867 & 0.610 & 2.543/2.413 & 31.268 & 1.762 & 19.866 & 0.626 & 2.910/2.937 & 41.200 & 4.445 \\
EDTalk~\cite{edtalk}    & 21.628&0.722 & 1.537/\textbf{1.290} & 17.698 & 8.115 & 25.156 & 0.811 & 1.676/1.315 & 13.785 & 7.642  \\
PortraitTalk~\cite{2024portraittalk} & 23.097 & 0.873 & 1.206/1.385 & 17.351 & 8.916 & 27.495 & 0.846 & 1.157/1.017 & 11.753 & 8.381 \\
\textbf{MAGIC-Talk} & \textbf{23.162} & \textbf{0.879} & \textbf{1.194}/1.368 & \textbf{17.236} & \textbf{8.958} & \textbf{ 27.563} & \textbf{0.892} & \textbf{1.126}/\textbf{1.009} & \textbf{11.671}  & 8.429 \\
\bottomrule
\end{tabular}
}

\caption{Quantitative comparison of MAGIC-Talk. The best-performing results are highlighted in bold. Arrows (↑ and ↓) indicate whether higher or lower values are preferable for each metric.} 
\label{table1}
\end{table*}

\subsection{Quantitative Results}
As shown in Table~\ref{table1}, we compare MAGIC-talk with state-of-the-art audio-to-talking face generation approaches on the HDTF and MEAD datasets. Our framework achieves superior identity preservation with higher PSNR, FID, and SSIM scores and demonstrates strong audio-lip synchronization and temporal consistency, as indicated by the high LMD and SyncNet scores. While Wav2Lip achieves the highest SyncNet score on HDTF due to using SyncNet as a training loss, MAGIC-talk ranks second on HDTF and achieves the highest score on MEAD, highlighting its effectiveness in audio-lip alignment. EDTalk reports higher LMD scores on MEAD since it leverages emotion videos as references for expressive face generation, whereas our framework relies on text prompts, making it inherently challenging to match the exact emotional expressions implicit in the ground truth text. Nevertheless, our framework consistently outperforms EDTalk across all other metrics, demonstrating its robustness in generating expressive, identity-consistent talking faces.

\subsection{Qualitative Results}
We compare MAGIC-talk with state-of-the-art methods, as shown in Figure~\ref{fig3}. The early methods like MakeItTalk and Wav2Lip suffer from artifacts and limited realism, with Wav2Lip introducing noticeable distortions due to its focus on modifying only the lip region. Methods such as Audio2Head, SadTalker, and TalkLip struggle with audio-lip synchronization, often generating restricted lip movements or unnatural closed-mouth faces. IP-LAP further fails to maintain synchronization, particularly for unseen identities. 
Emotion-driven methods like EMOGen, EAMM, and EDTalk introduce expressions but face challenges with identity preservation and motion consistency. EMOGen and EAMM produce inconsistent expressions and blurry artifacts, while EDTalk improves expression quality but struggles with natural head and shoulder movements. PortraitTalk, a prompt-driven model, achieves a better identity preservation but relies on multiple reference images, limiting its practicality. It also struggles with fine-grained emotional expressions and maintaining temporal consistency, often leading to misaligned head and hair movements. 
In contrast, our framework generates realistic expressions, precise audio-lip synchronization, and temporally coherent videos, all while requiring only a single reference image. This demonstrates superior robustness and practicality for real-world applications.

\begin{figure}[!t]
    \centering
    \includegraphics[trim={1.2cm 11cm 1.5cm 1cm}, clip, width=0.9\columnwidth]{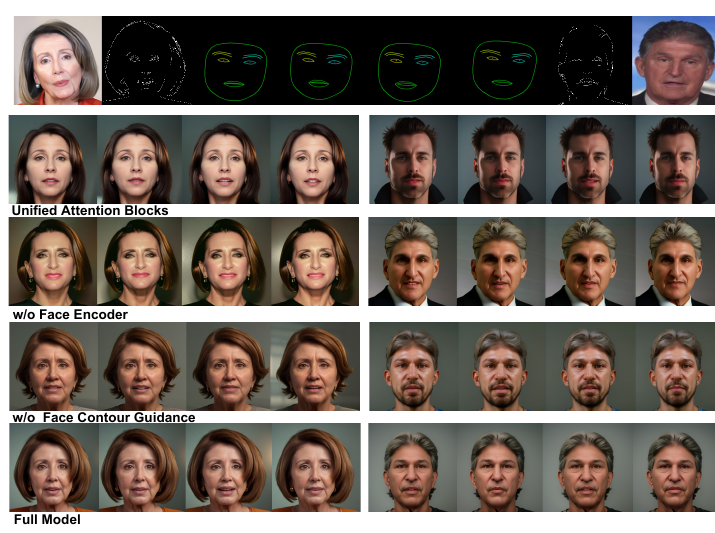}
    \caption{Illustration of the ablation study. Depicting the impact of key components in MAGIC-Talk.}
    \label{fig4}
\end{figure}

\subsection{Ablation Study}

\begin{table}[!t]
    \centering
    \resizebox{\columnwidth}{!}{%
        \begin{tabular}{lcccc}
            \toprule
            \textbf{Metric/Method} & \textbf{PSNR↑} & \textbf{SSIM↑} & \textbf{FID↓} & \textbf{SyncNet↑}\\ \midrule
            Unified attention block &9.518 & 0.284 & 16.083 & 0.047 \\
            w/o Face Encoder & 13.846 & 0.527 & 12.869 & 2.961 \\
            w/o face Contour guidance & 26.204 & 0.625 & 12.471 & 7.358  \\
            \textbf{Full model} & 27.563 & 0.892 & 11.671 & 8.429 \\ 
            
            \bottomrule
        \end{tabular}
    }
    \caption{A quantitative ablation study, evaluating the impact of the key components of the MAGIC-Talk framework.}
    \label{table2}
\end{table}

\paragraph{Unified Attention Block}To assess the effectiveness of the decoupled cross-attention mechanism, we replaced it with a standard cross-attention approach. As illustrated in Table~\ref{table2} and Figure~\ref{fig4}, the unified attention mechanism struggles to capture fine-grained facial details and maintain a coherent facial motion. While it shows some ability to interpret textual prompts, such as recognizing the speaker's gender, it fails to effectively integrate multiple input conditions, resulting in outputs that lack realism and deviate from the intended identity.

\paragraph{Image Encoder}We assess the effectiveness of the face encoder by substituting it with the widely used CLIP image encoder. The results reveal a significant decline in the facial feature preservation and structural fidelity in the generated videos. This inconsistency leads to identities that lack coherence across frames, greatly reducing the realism and quality of the talking faces. These findings emphasize the critical role of a specialized face encoder in ensuring identity-consistent talking faces.

\paragraph{Without Facial Contour Guidance}Maintaining the identity accuracy and consistent facial details from only one reference image is critical yet challenging task in talking face video generation. Any deviation in identity representation across frames can undermine the realism and temporal coherence of the video. To address this, we examined the impact of face contour guidance in maintaining the overall facial structure. As illustrated in Figure~\ref{fig4}, excluding face contour guidance leads to visible deformations and a reduced similarity to the reference image’s identity attributes. Conversely, incorporating face contours enhances structural consistency, improving both the identity fidelity and realism. By providing a foundational structure, contours help the model preserve spatial relationships between facial features, ensuring that the details like the jawline and cheek structure remain consistent across frames.

\begin{figure}[!t]
    \centering
    \includegraphics[trim={2.8cm 5.7cm 1cm 1cm}, clip, width=\columnwidth]{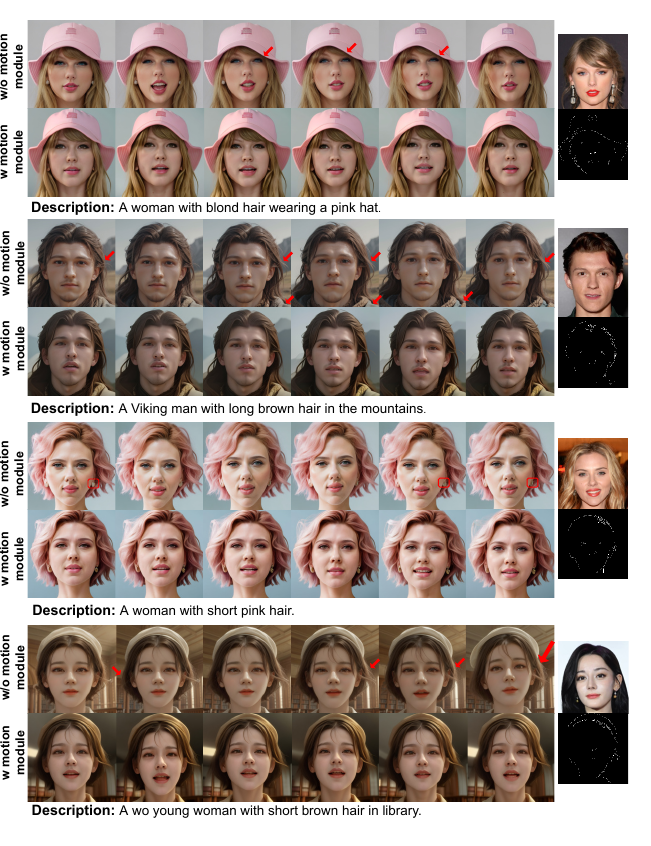}
    \caption{Effect of the motion module on talking face generation.    
    }
    \label{fig5}
\end{figure}

\subsection{Impact of Motion Module}
We investigate the effectiveness of incorporating a motion module to generate consistent talking faces. As shown in Figure~\ref{fig5}, integrating the motion module significantly enhances the smoothness and coherence of the generated videos. This improvement is particularly evident in the alignment of facial features and natural head movements, leading to more realistic and engaging animations. The quantitative results in Table \ref{table3} further validate these observations. The motion module improves both identity preservation and prompt fidelity, ensuring the generated video accurately reflects the user’s input, while maintaining the reference identity characteristics.

\begin{table}[!h]
    \centering
    \resizebox{\columnwidth}{!}{%
        \begin{tabular}{lcccc}
            \toprule
            \textbf{Metric/Method} & \textbf{CLIP-T\% ↑} & \textbf{DINO\% ↑} & \textbf{Face.sim\% ↑} & \textbf{SyncNet ↑}\\ \midrule
            w/o motion module & 21.409 & 76.3 & 72.9  & 5.258  \\
            \textbf{w motion module} & 21.412 & 77.4 & 73.6 & 6.416 \\

            \bottomrule
        \end{tabular}
    }
    \caption{A quantitative evaluation of the impact of the motion module on talking face generation. 
    }
    \label{table3}
\end{table}

\subsection{Long-form Video Generation}
\begin{figure}[!t]
    \centering
    \includegraphics[trim={1.1cm 8.1cm 1cm 1cm}, clip, width=\columnwidth]{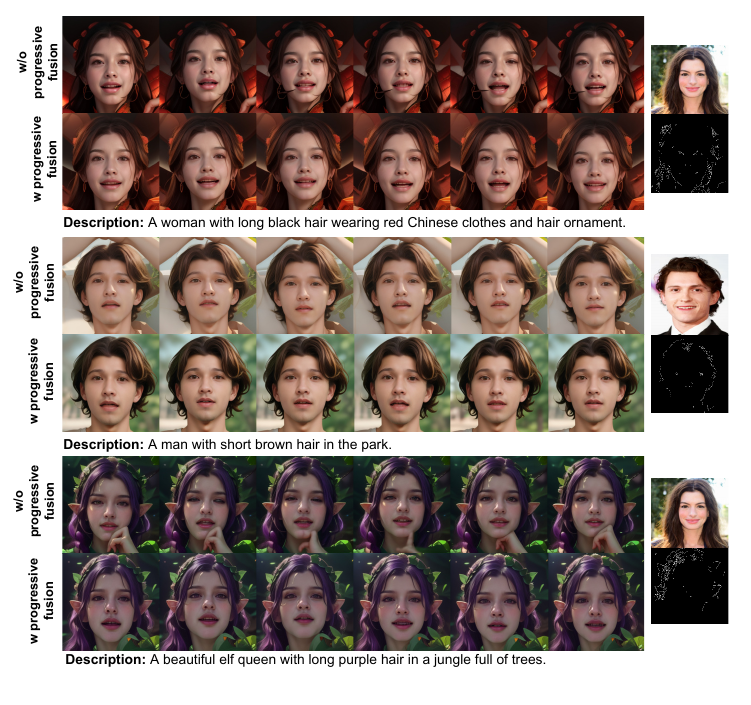}
    \caption{A comparison of the frames generated with and without the progressive fusion. 
    }
    \label{fig6}
\end{figure}

We evaluate the effectiveness of the progressive sampling fusion employed in one-shot talking face generation. 
As shown in Figure~\ref{fig6}, the progressive fusion significantly reduces artifacts, eliminates abrupt head movements, and improves lip synchronization. The weighted interpolation in overlapping frames ensures that temporal consistency is maintained without introducing noticeable blending artifacts. 
It is important to note that the frames shown in Figure~\ref{fig6} are not consecutive. They identify the frames where artifacts, incorrect lip movements, and unnatural head positioning occurred. 

\subsection{Expressive Talking Face Generation}
\begin{figure}[!t]
    \centering
    \includegraphics[trim={1.2cm 10.1cm 1.2cm 1.2cm}, clip, width=\columnwidth]{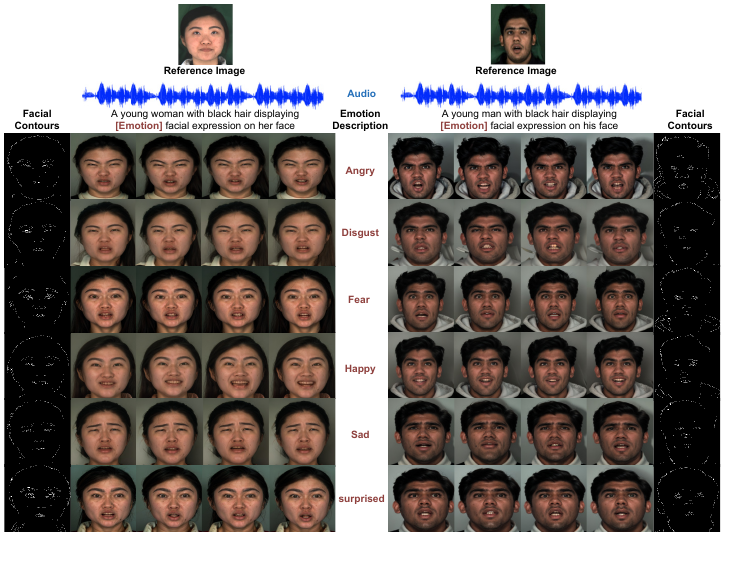}
    \caption{Expressive talking face generation with MAGIC-Talk, translating text-described emotions into realistic facial expressions with fine details.
    }
    \label{fig7}
\end{figure}

In this section, we evaluate the effectiveness of MAGIC-Talk in generating expressive talking faces using text descriptions. As shown in Figure~\ref{fig7}, MAGIC-Talk effectively translates the intended emotion from text descriptions into the generated talking faces while preserving identity and audio-lip synchronization. Our framework employs separate cross-attention for each conditioning input, enabling precise feature learning and providing better control over facial detail generation. Additionally, incorporating facial contours as guidance enhances structural consistency, resulting in more natural expressions. This leads to expressive and emotionally rich facial animations, capturing subtle details such as eyebrow movements, lip shaping, and overall facial dynamics. 

\section{Conclusion}
In this paper, we introduced MAGIC-Talk, a one-shot talking face generation framework that enables editable and audio-aligned talking faces. By integrating ReferenceNet and AnimateNet, our approach ensures customizable identity generation, while maintaining temporal consistency for long video synthesis. 
Extensive experiments and ablation studies demonstrate that MAGIC-Talk outperforms existing methods in portrait animation, achieving high-fidelity identity preservation, natural motion dynamics, and precise audio-lip synchronization. Our framework marks a significant advancement in controllable, generalizable, and temporally coherent talking face generation, making it well-suited for applications in virtual avatars, filmmaking, and digital content creation.




{
    \small
    \bibliographystyle{ieeenat_fullname}
    \bibliography{main}
}

\end{document}